\documentclass[11pt]{article}

\usepackage[preprint]{acl}

\usepackage{times}
\usepackage{latexsym}

\usepackage[T1]{fontenc}

\usepackage[utf8]{inputenc}

\usepackage{microtype}

\usepackage{inconsolata}

\usepackage{graphicx}

\usepackage{verbatim}       
\usepackage{amsmath}        
\usepackage{amssymb}        
\usepackage{algorithm}      
\usepackage{algpseudocode}  
\usepackage{tikz}           
\usetikzlibrary{calc,positioning,arrows.meta,fit}
\usepackage{tcolorbox}      
\tcbuselibrary{breakable,listings}   
\usepackage{float}          
\usepackage{placeins}       
\usepackage{subcaption}     
\usepackage{multirow}       
\usepackage{listings}       
\usepackage[table]{xcolor}  

\newcommand{\Rationale}[1]{\rho_{#1}}
\newcommand{\Code}[1]{\mathcal{A}_{#1}}
\newcommand{\Output}[1]{\mathcal{O}_{#1}}

\definecolor{panelgray}{RGB}{245,245,245}
\definecolor{pastelblue}{RGB}{223,228,250}
\definecolor{pastelpink}{RGB}{250,228,228}
\definecolor{pastelgreen}{RGB}{220,245,220}

\newcommand{\gapRP}{0.35cm}
\newcommand{\distInputLLM}{3cm}
\newcommand{\distLLMFT}{3cm}
\newcommand{\arrowPad}{0.20cm}

\tikzset{
  panel/.style = {
    draw,
    rounded corners=10pt,
    fill=panelgray,
    minimum width=#1,
    minimum height=3.4cm,
    line width=0.6pt
  },
  panel/.default=7.2cm,
  pill/.style = {
    draw,
    rounded corners=8pt,
    inner sep=7pt,
    font=\normalsize,
    minimum height=1.15cm,
    minimum width=2.8cm
  },
  pillPink/.style  = {pill, fill=pastelpink},
  pillGreen/.style = {pill, fill=pastelgreen},
  pillBlue/.style  = {pill, fill=pastelblue},
  arrow/.style     = {->, >=Stealth, line width=0.7pt},
  cycleArrow/.style= {->, >=Stealth, line width=0.85pt},
  trajArrow/.style = {->, >=Stealth, line width=0.75pt, shorten >=3pt, shorten <=3pt}
}

\title{VisTIRA: Closing the Image–Text Modality Gap in Visual Math Reasoning via Structured Tool Integration}

\author{Saeed Khaki \\
  Microsoft AI \\
  {\footnotesize \texttt{saeedkhaki@microsoft.com}} \\\And
  Ashudeep Singh \\
  Microsoft AI \\
  {\footnotesize \texttt{ashudeep.singh@microsoft.com}} \\\And
  Nima Safaei \\
  Ohio State University \\
  {\footnotesize \texttt{safaei.3@osu.edu}} \\\AND
  Kamal Ginotra \\
  Microsoft AI \\
  {\footnotesize \texttt{kamalginotra@microsoft.com}} \\}

\begin{document}
\maketitle
\begin{abstract}
Vision-language models (VLMs) lag behind text-only language models on mathematical reasoning when the same problems are presented as images rather than text. We empirically characterize this as a \textit{modality gap}: the same question in text form yields markedly higher accuracy than its visually typeset counterpart, due to compounded failures in reading dense formulas, layout, and mixed symbolic--diagrammatic context. First, we introduce VisTIRA (Vision and Tool-Integrated Reasoning Agent), a tool-integrated reasoning framework that enables structured problem solving by iteratively decomposing a given math problem (as an image) into natural language rationales and executable Python steps to determine the final answer. Second, we build a framework to measure and improve visual math reasoning: a LaTeX-based pipeline that converts chain-of-thought math corpora (e.g., NuminaMath) into challenging image counterparts, and a large set of synthetic tool-use trajectories derived from a real-world, homework-style image dataset (called SnapAsk) for fine-tuning VLMs. Our experiments show that tool-integrated supervision improves image-based reasoning, and OCR grounding can further narrow the gap for smaller models, although its benefit diminishes at scale. These findings highlight that modality gap severity inversely correlates with model size, and that structured reasoning and OCR-based grounding are complementary strategies for advancing visual mathematical reasoning.
\end{abstract}

\section{Introduction}

Vision-language models (VLMs) have achieved strong performance on generic multimodal tasks, including document understanding and visual question answering (e.g., DocVQA and chart- or plot-based QA benchmarks) \citep{mathew2021docvqa,masry2022chartqa,methani2020plotqa}. However, when mathematical problems are presented as images that mix dense symbolic expressions, multi-line equations, diagrams, plots, and embedded textual instructions, current VLMs still trail their text-only large language model counterparts. This discrepancy manifests as a persistent \emph{image-text modality gap}: the same math question, when rendered visually rather than provided as plaintext, yields significantly lower accuracy, as shown by recent multimodal math benchmarks \citep{lu2024mathvista,wang2024mathv,wang2025vcbench,zhang2024mathverse,wang2025mvmath} and analyses that formalize modality and capability mismatches in VLMs \citep{yi2025swab,schrodi2025modalitygap}. Even frontier systems show degradation on visual math, whereas mid-sized open models suffer large drops \citep{lu2024mathvista,wang2024mathv}. Accurate mathematical reasoning in the visual setting requires two coupled competencies: (i) faithful visual parsing of layout-level structure, such as fraction baselines, integral bounds, superscripts and subscripts, piecewise braces, and diagram annotations, and (ii) robust symbolic and quantitative reasoning \citep{wang2024unimernet,zhong2025doctronformula,blecher2024nougat}. Small misreads cascade; for example, an incorrectly transcribed exponent or reversed limit alters subsequent algebra, leading to compounding errors until the final answer diverges. Similar cascades have been documented in visual numeracy tasks, where mis-parsed axes or legends propagate into incorrect computations \citep{masry2022chartqa,methani2020plotqa}.

Recent advances in textual mathematical reasoning have leveraged chain-of-thought (CoT) prompting \citep{wei2022cot}, program-of-thought (PoT) execution, and tool integration (e.g., symbolic solvers and Python libraries) \citep{gao2022pal}, yielding substantial gains on math and symbolic tasks \citep{sprague2025tocot,chen2023pot}. In contrast, visual mathematical reasoning has received less attention. Existing VQA and OCR-style datasets rarely contain stepwise executable trajectories paired with images and typically provide only final answers or short natural language rationales \citep{lu2024mathvista,masry2022chartqa,methani2020plotqa}. Bridging this gap requires frameworks and data that align visual perception with iterative computation.

We introduce VisTIRA (Vision and Tool-Integrated Reasoning Agent), a framework that equips a VLM to iteratively solve image-based math problems by interleaving natural language rationales with executable Python programs \citep{suris2023vipergpt}. At each step, the model proposes a rationale and a code snippet; the code is executed externally, its output is appended to the evolving trajectory, and the model decides whether to stop after producing a boxed final answer or to continue refining. This closed-loop design reduces hallucinated arithmetic by deferring exact manipulation to a symbolic engine and by providing verifiable intermediate signals, consistent with evidence that program execution mitigates reasoning errors \citep{chen2023pot,gou2023tora} and with broader VLM hallucination mitigation via verification-style signals \citep{zhang2024vluncertainty,wu2025reverse,park2025halloc,sahu2024pelican}.

To supervise VisTIRA, we construct a large corpus of high-quality tool-integrated trajectories from real-world homework-style images (SnapAsk), generated by prompting strong teacher models and filtered for internal consistency \citep{chen-etal-2024-measuring}. Our data construction strategy is informed by prior trajectory-supervised tool-augmented math corpora in the text domain \citep{gou2023tora,zhang2023carp}, but extends them to the visual domain with explicit layout-aware parsing and executable traces. Separately, we develop a LaTeX-based text-to-image rendering pipeline that converts existing text-only chain-of-thought math problems, such as NuminaMath, into visually typeset images, enabling controlled modality gap evaluation \citep{numina2024numinamath,skripkin2025renderedtext}. Using this pipeline, we generate and release 360k rendered NuminaMath images to advance open research in mathematical visual reasoning.

We also investigate OCR grounding as a complementary lever for mitigating the modality gap. Applying a state-of-the-art OCR system, such as DeepSeek-OCR \citep{wei2025deepseek}, to extract textual content from math images and then inputting it alongside the original image markedly improves accuracy for smaller VLMs \citep{shenoy2024lumos}, indicating that explicit text extraction can compensate for weaker visual encoders, in line with findings from document VQA \citep{mathew2021docvqa} and recent OCR advances \citep{wei2025deepseek}. For larger models, however, raw OCR concatenation can introduce redundancy or noise, revealing a scale-dependent trade-off \citep{baek2025ocrheads} consistent with observations in large VLM hallucination analyses \citep{zhang2024vluncertainty,wu2025reverse}.

Comprehensive experiments on the real-world SnapAsk and rendered NuminaMath benchmarks show that supervised fine-tuning on VisTIRA trajectories yields measurable gains over instruction-only baselines \citep{gou2023tora,shi2024mathllava}. They also show that the modality gap remains substantial, particularly for smaller models, but can be partially mitigated via tool-integrated reasoning and OCR-based grounding \citep{lu2024mathvista,wang2024mathv,wang2025vcbench,zhang2024mathverse}.

\paragraph{Contributions.}
We summarize our main contributions:\vspace{-0.4em}
\begin{itemize}
  \item \textbf{Framework:} We propose VisTIRA, an iterative tool-integrated reasoning framework that decomposes visual math problems into rationale, code, and execution loops until solution \citep{chen2023pot,gou2023critic, suris2023vipergpt}.
  \item \textbf{Trajectory Supervision:} We construct a large corpus of verified rationale--code--output trajectories from real-world homework-style images (SnapAsk), enabling supervised fine-tuning of VLM math agents in the visual setting \citep{gou2023tora,zhang2023carp}.
  \item \textbf{Evaluation Pipeline \& Data Release:} We introduce a LaTeX-based rendering pipeline to convert text-only CoT math corpora (e.g., NuminaMath) into paired image modalities for controlled modality gap analysis, and release 360k clean and 360k noisy rendered images generated via a stochastic degradation pipeline \citep{numina2024numinamath,skripkin2025renderedtext}. We also release a 5k NuminaMath image test set with DeepSeek-OCR transcriptions as a standardized benchmark \citep{wei2025deepseek}.
  \item \textbf{OCR Grounding Analysis:} We present a diagnostic analysis showing that OCR text extraction (DeepSeek-OCR) provides substantial gains for smaller models but exhibits scale-dependent diminishing returns, motivating future work on adaptive fusion \citep{mathew2021docvqa,wei2025deepseek,baek2025ocrheads}.
  \item \textbf{Empirical Findings:} We quantify the image--text modality gap across model scales and demonstrate partial mitigation via tool-integrated reasoning plus OCR grounding, aligning with trends reported in recent multimodal math benchmarks \citep{lu2024mathvista,wang2024mathv,wang2025vcbench,zhang2024mathverse,wang2025mvmath}.
\end{itemize}

Figure~\ref{fig:workflow} summarizes our workflow: we generate verified tool-integrated trajectories from SnapAsk for supervised fine-tuning, render a paired NuminaMath text$\leftrightarrow$image benchmark via LaTeX for controlled modality-gap analysis, and evaluate the resulting VisTIRA model on both SnapAsk and rendered NuminaMath, optionally with OCR grounding.

\begin{figure*}[t]
\centering
\resizebox{\textwidth}{!}{%
\begin{tikzpicture}[
    box/.style={
        draw, rectangle, rounded corners=3pt,
        minimum width=4.8cm, minimum height=1.4cm,
        text width=4.6cm, align=center,
        font=\small, inner sep=6pt,
        fill=white, line width=0.5pt
    },
    boxB/.style={
        draw, rectangle, rounded corners=3pt,
        minimum width=5.4cm, minimum height=1.4cm,
        text width=5.2cm, align=center,
        font=\small, inner sep=6pt,
        fill=white, line width=0.5pt
    },
    boxC/.style={
        draw, rectangle, rounded corners=3pt,
        minimum width=5.0cm, minimum height=1.4cm,
        text width=4.8cm, align=center,
        font=\small, inner sep=6pt,
        fill=white, line width=0.5pt
    },
    evalbox/.style={
        draw, rectangle, rounded corners=3pt,
        minimum height=2.2cm,
        text width=14.5cm, align=left,
        font=\small, inner sep=8pt,
        fill=white, line width=0.5pt
    },
    colhead/.style={font=\normalsize\bfseries},
    arr/.style={->, >=Stealth, line width=0.6pt},
    darr/.style={->, >=Stealth, line width=0.6pt, dashed},
    every node/.append style={font=\small}
]

\def\colA{0}
\def\colB{7.5}
\def\colC{15}

\def\rowTitle{0}
\def\rowOne{-1.8}
\def\rowTwo{-4.0}
\def\rowThree{-6.2}
\def\rowFour{-8.6}
\def\rowFive{-11.0}
\def\rowEval{-14.0}

\node[colhead] at (\colA, \rowTitle) {\textbf{A. Training supervision (SnapAsk)}};

\node[box] (A1) at (\colA, \rowOne) {
    SnapAsk (train)\\
    {\footnotesize (real homework-style}\\
    {\footnotesize math images)}
};

\node[box] (A2) at (\colA, \rowTwo) {
    Teacher VLM\\
    {\footnotesize (rationale $\rightarrow$ code)}\\
    {\footnotesize + external tool execution}\\
    {\footnotesize (Python/SymPy)}
};

\node[box] (A3) at (\colA, \rowThree) {
    Tool-integrated trajectories\\
    {\footnotesize $(\rho_1,\!\mathcal{A}_1,\!\mathcal{O}_1,\!\rho_2,\!\mathcal{A}_2,\!\mathcal{O}_2,\!\ldots,\!\rho_n,\!\mathcal{A}_n,\!\mathcal{O}_n)$}\\
    {\footnotesize + final \texttt{\textbackslash boxed\{answer\}}}
};

\node[box] (A4) at (\colA, \rowFour) {
    Consistency filter\\
    {\footnotesize (keep trajectories where}\\
    {\footnotesize CoT answer matches}\\
    {\footnotesize final \texttt{\textbackslash boxed\{answer\}})}
};

\node[box] (A5) at (\colA, \rowFive) {
    SnapAsk (test)\\
    {\footnotesize (held-out 5k images)}\\
    {\footnotesize for real-world eval}
};

\draw[arr] (A1) -- (A2);
\draw[arr] (A2) -- (A3);
\draw[arr] (A3) -- (A4);
\draw[arr] (A4) -- (A5);

\node[colhead] at (\colB, \rowTitle) {\textbf{B. Train VisTIRA agent}};

\node[boxB] (B1) at (\colB, \rowOne) {
    Base VLM\\
    {\footnotesize (e.g., Qwen2.5-VL-7B-Instruct)}
};

\node[boxB] (B2) at (\colB, \rowThree) {
    SFT on verified trajectories\\
    {\footnotesize $\rightarrow$ VisTIRA-finetuned policy}
};

\node[boxB] (B3) at (\colB, \rowFour) {
    VisTIRA agent at inference\\
    {\footnotesize (finetuned VLM +}\\
    {\footnotesize external tool executor)}
};

\draw[arr] (B1) -- (B2);
\draw[arr] (B2) -- (B3);

\node[colhead] at (\colC, \rowTitle) {\textbf{C. Paired benchmark + OCR (NuminaMath)}};

\node[boxC] (C1) at (\colC, \rowOne) {
    NuminaMath-CoT (text)\\
    {\footnotesize (problem + CoT + answer)}
};

\node[boxC] (C2) at (\colC, \rowTwo) {
    LLM conversion\\
    {\footnotesize (structure-preserving rules)}\\
    {\footnotesize $\rightarrow$ compilable LaTeX}
};

\node[boxC] (C3) at (\colC, \rowThree) {
    LaTeX compile + render\\
    {\footnotesize $\rightarrow$ typeset problem images}\\
    {\footnotesize (preserve original answers)}
};

\node[boxC] (C4) at (\colC, \rowFour) {
    OCR\\
    {\footnotesize (DeepSeek-OCR)}\\
    {\footnotesize (optional grounding)}\\
    {\footnotesize for rendered benchmark}
};

\draw[arr] (C1) -- (C2);
\draw[arr] (C2) -- (C3);
\draw[arr] (C3) -- (C4);


\draw[arr] (A4.east) -- ++(0.3, 0) |- (B2.west);
\node[font=\footnotesize\itshape, fill=white, inner sep=2pt] at ($(A4.east)!0.5!(B2.west) + (0, -0.55)$) {VisTIRA-SFT set};

\node[evalbox] (eval) at (\colB, \rowEval) {
    \hspace{1.2cm}\underline{\textbf{Evaluation}}\\[4pt]
    \hspace{0.2cm}$\bullet$\;\; Rendered NuminaMath (paired): text vs image vs image+OCR vs OCR-only $\rightarrow$ modality gap\\
    \hspace{0.2cm}$\bullet$\;\; SnapAsk (real): image input $\rightarrow$ real-world accuracy\\
    \hspace{0.2cm}$\bullet$\;\; Report gains from tool-integrated supervision (VisTIRA) and OCR grounding
};

\coordinate (A5mid) at ($(A5.south) + (0, -0.7)$);
\coordinate (A5eval) at ([xshift=-5.5cm]eval.north);
\draw[arr] (A5.south) -- (A5mid) -- (A5mid -| A5eval) -- (A5eval);
\draw[arr] (B3.south) -- (eval.north);
\coordinate (C4mid) at ($(C4.south) + (0, -0.7)$);
\coordinate (C4eval) at ([xshift=5.5cm]eval.north);
\draw[arr] (C4.south) -- (C4mid) -- (C4mid -| C4eval) -- (C4eval);

\end{tikzpicture}
}
\caption{End-to-End VisTIRA Workflow. (A) From SnapAsk math images, a teacher VLM generates tool-integrated trajectories (rationale + executable Python with external execution), which are filtered via answer-consistency and used for SFT. (B) We construct a paired text$\leftrightarrow$image benchmark by converting NuminaMath problems to compilable LaTeX and rendering typeset images while preserving original answers. (C) We fine-tune a base VLM on the verified trajectories and evaluate on SnapAsk and rendered NuminaMath to measure the text--image modality gap, optionally adding OCR grounding.}
\label{fig:workflow}
\end{figure*}

\section{Method}

\subsection{Vision--Language Models as Tool-Integrated Math Agents}

To address the modality gap described above, we propose a framework that augments vision-language models with tool-integrated reasoning, combining natural language inference with computational engines to bridge interpretive capabilities and mathematical precision.

Our proposed Vision and Tool-Integrated Reasoning Agent (VisTIRA) approaches a visual mathematical problem, represented as an image $I$, by decomposing it into a sequence of natural language reasoning steps, denoted as $\rho_i$, and corresponding tool-based actions, denoted as $\mathcal{A}_i$, such as free-form symbolic Python programs. This structured decomposition enables VisTIRA to combine interpretive guidance with computational execution to solve math problems presented in visual formats. At each step, the program $\mathcal{A}_i$ is executed and its output $\mathcal{O}_i$ is fed back into VisTIRA \cite{suris2023vipergpt}. This output informs the next stage of processing, which may involve generating new reasoning steps, refining the program, or preparing the final answer. This iterative loop enables VisTIRA to adaptively solve visual math problems through a combination of reasoning and tool-based computation. The process iterates until the model's output includes the final answer enclosed in \verb|\boxed{}| \cite{gou2023tora}. The generated reasoning trajectory is formalized in Equation~\ref{eq:VisTIRA}, and the overall VisTIRA data generation process is visualized in Figure~\ref{fig:diagram-VisTIRA}. Additional trajectory examples are provided in Appendix~\ref{appendix:VisTIRA-examples}.

To construct this dataset, we prompt advanced vision-language models (e.g., GPT-5 \cite{openai2025gpt5}, Gemini \cite{gemini2025}) to produce tool-integrated reasoning sequences, which we term VisTIRA data. This synthetic corpus is then used to fine-tune a smaller VLM, equipping it to serve as a mathematical reasoning agent.

\subsubsection{VisTIRA Data Generation}

Existing training datasets for vision-language models are largely limited to tasks such as VQA or OCR, which do not encompass complex mathematical reasoning \citep{mathew2021docvqa,masry2022chartqa}. Moreover, these datasets typically provide only natural language annotations, lacking the structured, step-by-step tool-use supervision required to train tool-integrated reasoning agents \citep{gou2023tora,zhang2023carp}. To address this limitation, we generated high-quality, tool-integrated reasoning trajectories using a large proprietary dataset of real-world mathematical problems, known as SnapAsk. This dataset spans multiple domains, including mathematics, algebra, geometry, and physics, and covers a wide range of difficulty levels. Representative examples are shown in Figure~\ref{fig:hm_problems}.

\begin{algorithm}[t]
\caption{VisTIRA Inference Procedure. $\oplus$ denotes string concatenation.}
\label{alg:tool_inference}
\begin{algorithmic}[1]
\Require Math Image $I$, model $\mathcal{M}$, base prompt $p$, external tools $\mathcal{U}$, stopping rule $\mathrm{Stop}(\cdot)$, and maximum steps $k$
\State $\tau_0 \gets ``\:"$ \Comment{Initialize reasoning trajectory}
\For{$i \gets 1$ \textbf{to} $k$}
    \State $\rho_i \sim \mathbb{P}_{\mathcal{M}}(\cdot \mid p \oplus I \oplus \tau_{i-1})$
    \If{$\mathrm{Stop}(\rho_i)$}
        \State \Return $\tau_{i-1} \oplus \rho_i$ \Comment{Stopping criterion met}
    \EndIf
    \State $\mathcal{A}_i \sim \mathbb{P}_{\mathcal{M}}(\cdot \mid p \oplus I \oplus \tau_{i-1} \oplus \rho_i)$
    \State $\mathcal{O}_i \gets \mathcal{U}(\mathcal{A}_i)$ \Comment{Tool execution}
    \State $\tau_i \gets \tau_{i-1} \oplus \rho_i \oplus \mathcal{A}_i \oplus \mathcal{O}_i$
\EndFor
\State \Return $\tau_k$
\end{algorithmic}
\end{algorithm}

\begin{equation}
\tau = \Rationale{1} \Code{1} \Output{1} \ldots \rho_{n-1} \mathcal{A}_{n-1} \mathcal{O}_{n-1} \rho_n
\label{eq:VisTIRA}
\end{equation}

Algorithm~\ref{alg:tool_inference} presents the step-by-step inference process for generating VisTIRA trajectories, which include both natural language rationales and executable programs. The process begins with a detailed prompt ($p$), enriched with diverse few-shot examples (see Appendix~\ref{appendix:VisTIRA}), that guides a large vision-language model (e.g., GPT-5 \cite{openai2025gpt5}) to decompose the visual problem into a sequence of rationales and corresponding Python programs. Each generated program is executed, and its output is fed back to the model in the next step. When the model emits a designated code-execution trigger, such as the stop word ``\verb|```output|'', we execute the corresponding program and return its output $O_i$ by invoking the tool as $O_i \leftarrow \mathcal{U}(A_i)$. This output is then provided to the model to facilitate the generation of subsequent reasoning steps \cite{gou2023critic}. The process continues iteratively until a stopping condition is met, such as the appearance of a final answer enclosed in \verb|\boxed{}|, at which point the trajectory is complete. Appendix~\ref{appendix:VisTIRA-examples} presents sample VisTIRA trajectories generated by GPT-4o \cite{openai2024gpt4o}.

\subsubsection{Supervised Fine-Tuning (SFT)}

We perform supervised fine-tuning on the high-quality VisTIRA dataset generated using large vision-language models. This dataset is denoted as $D_{\text{sft}} = \{(I_1, x_1, y_1), (I_2, x_2, y_2), \dots, (I_n, x_n, y_n)\}$~\cite{khaki2024rsdpo,ouyang2022training}, where $I_i$ represents the $i$-th image, $x_i$ the input prompt, and $y_i$ the corresponding VisTIRA response. Starting from a base VLM, supervised fine-tuning maximizes the likelihood of generating the response $y$ conditioned on the prompt $x$ and image $I$, as formalized in Equation~\ref{eq:sft}.

\begin{equation}
\mathcal{L}^{\text{SFT}} = \arg\max \sum_{(I,x,y) \in \mathcal{D}_{\text{sft}}} \log \pi(y \mid I,x)
\label{eq:sft}
\end{equation}

\subsection{LaTeX-Based Text to Image Conversion}

To address the scarcity of image--text supervision for mathematical reasoning, we synthesize such data from text-only corpora via LaTeX-based rendering \citep{skripkin2025renderedtext,yamabe2025tpi}. Specifically, we convert text-only math problems into visually typeset inputs while preserving their original textual answers, yielding image-to-text data pairs that align training with realistic deployment conditions (e.g., homework sheets and exams where questions and figures co-occur).

Our proposed method consists of the following steps. First, we design a detailed prompt for large language models (e.g., GPT-4o, Claude \cite{claude_personal_2025}) that specifies how to convert a math question (along with an associated image, if present) into compilable LaTeX code that preserves all mathematical notations and structures (see Appendix~\ref{appendix:text-to-latex}). Next, we compile the generated LaTeX code with a LaTeX engine to produce a PDF \citep{skripkin2025renderedtext}, which we then render into an image modality to serve as input for VLM evaluation or training. The response modality remains text, as it is the standard output format for VLMs. Appendix~\ref{appendix:text_to_image_examples} presents examples of mathematical problems in text modality, the corresponding LaTeX code generated by GPT-4o, and the rendered image output. This compilation-and-rendering process enables direct comparison between textual and visual representations of the same problem. As demonstrated in these examples, the text and image modalities differ substantially, with the image modality presenting a greater challenge for vision-language models in accurately understanding and solving mathematical problems. This increased difficulty arises from the need to visually interpret complex structures such as equations, diagrams, and embedded text. We refer to this performance discrepancy as the modality gap, which is a central focus of our analysis in this paper. The complete pipeline is illustrated in Figure~\ref{fig:pipeline-eval-gen}.

\section{Experimental Details}

This section presents our experimental configuration and findings, highlighting the effectiveness of the proposed VisTIRA data generation pipeline for training vision-language models and evaluating its quality on both LaTeX-rendered and real-world mathematical image datasets. All experiments utilize Qwen2.5-VL-7B~\cite{qwen2.5,Qwen2VL,Qwen-VL}, a state-of-the-art VLM with 7B parameters. We perform supervised fine-tuning (SFT) on the generated dataset using DeepSpeed ZeRO-3~\cite{ren2021zero,rajbhandari2020zero,rasley2020deepspeed} to optimize memory usage and training efficiency. Training is carried out on 8 NVIDIA V100 GPUs (32 \,GB each).

For SFT, we adopt a cosine learning rate schedule with an initial learning rate of $2\times10^{-5}$, an effective batch size of 64, a single epoch, weight decay of 0.1, and a maximum sequence length of 8{,}192 tokens. Additionally, LoRA-based fine-tuning~\cite{hu2022lora} is applied with rank $r=32$, $\alpha=64$, and a dropout rate of 0.05.

\subsection{Datasets}

We use the following datasets in our experiments:

\textbf{SnapAsk:} SnapAsk is a large proprietary dataset of real-world mathematical problems with approximately 303k samples. The dataset spans multiple domains, including mathematics, algebra, geometry, and physics, and covers a wide range of difficulty levels. Representative examples are shown in Figure~\ref{fig:hm_problems}. This dataset is used to generate the VisTIRA training set to teach VLMs tool-integrated reasoning for solving complex mathematical problems. For the generation of the VisTIRA responses, we use GPT-4o \cite{openai2024gpt4o} as the teacher model in Algorithm~\ref{alg:tool_inference} with $n=2$ and greedy decoding to produce VisTIRA trajectories. Since the dataset does not contain ground-truth answers, we retain only images for which the model's chain-of-thought (CoT) response matches the final answer in the VisTIRA trajectory. After this verification step, we obtain 147{,}948 samples for VLM training and keep a hold-out test set of 5k images for the final evaluation.

\textbf{NuminaMath-CoT:} The NuminaMath dataset~\cite{numina2024numinamath} contains approximately 860k text-to-text math problems, with solutions expressed in a Chain-of-Thought (CoT) format. These problems range from standard exercises to Olympiad-level questions. From this dataset, we sample about 5k problems and convert them into a VQA format using our pipeline (Figure~\ref{fig:pipeline-eval-gen}) for evaluating VLMs \citep{skripkin2025renderedtext}. Appendix~\ref{appendix:text_to_image_examples} provides several examples of this conversion. We select this dataset for two main reasons: (1) it is a well-established math dataset with diverse difficulty levels, making it suitable for assessing VLM performance; and (2) using the original text version before conversion to image modality allows us to demonstrate the modality gap in VLM performance on the same set of questions.

Table~\ref{tab:dataset-summary} provides a compact summary of all datasets and splits used in this work. Additional trajectory-level statistics and representative examples are provided in Appendix~\ref{appendix:dataset-stats}.

\begin{table}[t]
\centering
\renewcommand{\arraystretch}{1.15}
\resizebox{\columnwidth}{!}{%
\begin{tabular}{llrl}
\hline
\textbf{Dataset} & \textbf{Split / Variant} & \textbf{Size} & \textbf{Purpose} \\
\hline
SnapAsk (raw) & Full corpus & $\sim$303k & Source images \\
SnapAsk (VisTIRA) & Train & 147{,}948 & SFT (tool-integrated) \\
SnapAsk (VisTIRA) & Test & 5{,}000 & Evaluation \\
NuminaMath-CoT & Source (text-only) & $\sim$860k & Text-to-image rendering \\
NuminaMath (rendered) & Clean images & 360k & Released for community \\
NuminaMath (rendered) & Noisy images & 360k & Released for community \\
NuminaMath (rendered) & Eval subset & 5{,}000 & Modality-gap evaluation \\
\hline
\end{tabular}}
\vspace{0.5em}
\caption{Summary of datasets and splits. SnapAsk provides real-world homework images; NuminaMath provides controlled text--image pairs. VisTIRA trajectories are retained only when the CoT answer matches the tool-execution answer (rate $\approx$48.9\%). See Appendix~\ref{appendix:dataset-stats}.}
\label{tab:dataset-summary}
\end{table}

\section{Evaluation and Results}

We design a comprehensive evaluation framework to address two key objectives: (1) assess the capability of vision-language models to perform tool-integrated reasoning on mathematical images, and (2) analyze the modality gap between text and image representations in the mathematical domain.

\subsection{Evaluating Tool-Integrated Reasoning in VLMs}

To evaluate models' ability to perform mathematical reasoning in visual contexts, we conduct experiments on two datasets: NuminaMath and SnapAsk, each containing 5{,}000 images. NuminaMath is a synthetic dataset generated using our proposed text-to-image LaTeX conversion pipeline. In contrast, SnapAsk consists of real-world user-uploaded mathematical images, enabling us to assess model performance in practical scenarios. Tables~\ref{tab:numinamath-comparison} and~\ref{tab:snapask-comparison} show model comparisons on the NuminaMath and SnapAsk datasets, respectively. The results indicate that Qwen2.5-VL-7B \citep{bai2025qwen25vl} successfully learns tool-integrated reasoning after being fine-tuned on the VisTIRA training corpus. This capability enables the model to tackle more complex mathematical problems for which traditional chain-of-thought reasoning alone may be insufficient \citep{chen2023pot,gou2023tora}. Appendix~\ref{app:vistira} presents detailed comparison examples illustrating how the fine-tuned Qwen2.5-VL-7B-VisTIRA model corrects errors made by the base model through tool-integrated reasoning.

\begin{table}[t]
\centering
\resizebox{\columnwidth}{!}{%
\begin{tabular}{lcc}
\hline
\textbf{Model} & \textbf{NuminaMath (\%)} & \textbf{SnapAsk (\%)} \\
\hline
Qwen2.5-VL-7B-Instruct & 58.77 & 32.53 \\
Qwen2.5-VL-7B-Instruct (CoT-only SFT) & 58.10 & 34.18 \\
Qwen2.5-VL-7B-VisTIRA & 60.97 & 37.96 \\
GPT-5 (CoT-only) & 73.94 & -- \\
\hline
\end{tabular}}
\vspace{0.5em}
\caption{Model comparisons on NuminaMath (5k rendered images) and SnapAsk (5k real-world images). GPT-5 serves as a CoT-only baseline (no tool integration). Qwen2.5-VL-7B-VisTIRA is fine-tuned on our VisTIRA corpus. CoT-only SFT is trained on the same data with standard CoT targets (no tool-use/code/execution), isolating domain adaptation from tool-integrated supervision. On SnapAsk, CoT-only SFT provides +1.65 points over the base, while VisTIRA provides +5.43 points.}
\label{tab:numinamath-comparison}
\label{tab:snapask-comparison}
\end{table}

\paragraph{CoT-Only Ablation Analysis.}
To disentangle the effect of domain adaptation from tool-integrated supervision, we trained Qwen2.5-VL-7B-Instruct on the exact same SnapAsk-derived training set using standard chain-of-thought (CoT) targets only---no tool-use, code generation, or execution. On NuminaMath, CoT-only SFT achieves 58.10\%, slightly below the base model (58.77\%), while VisTIRA reaches 60.97\%. On SnapAsk, CoT-only SFT achieves 34.18\% (+1.65 over base), whereas VisTIRA achieves 37.96\% (+5.43 over base). These results confirm that VisTIRA's gains are not explained by domain adaptation alone; the tool-integrated supervision---interleaving rationale, code, and execution---provides a distinct and substantial benefit, particularly on real-world images where symbolic computation reduces compounding errors.

\subsection{Analyzing Modality Gaps in Mathematical Problem Solving}

\begin{table*}[t]
\centering
\small
\renewcommand{\arraystretch}{1.15}
\begin{tabular}{lccc}
\hline
\textbf{Model} & \textbf{Input Modality Type} & \textbf{Accuracy (\%)} & \textbf{Notes} \\
\hline
\multirow{5}{*}{Qwen2.5-VL-7B-Instruct} 
  & Text Modality & 64.19 & -- \\
  & Image Modality & 58.77 & -- \\
  & Image Modality + VisTIRA & 60.97 & Fine-tuned on VisTIRA corpus \\
  & Image + OCR Grounding & 63.89 & OCR used as extra grounding \\
  & OCR-only Grounding & 58.80 & OCR used without image input \\
\hline
\multirow{4}{*}{GPT-5} 
  & Text Modality & 79.94 & -- \\
  & Image Modality & 73.94 & -- \\
  & Image + OCR Grounding & 76.37 & OCR used as extra grounding \\
  & OCR-only Grounding & 77.49 & OCR used without image input \\
\hline
\end{tabular}
\vspace{0.5em}
\caption{Performance comparison of Qwen2.5-VL-7B-Instruct and GPT-5 on the NuminaMath dataset across four input modalities: text, image, image with OCR grounding, and OCR-only. Results show that OCR-only input can outperform raw image modality, especially for Qwen, suggesting that textual grounding extracted from images provides stronger semantic cues than visual features alone.}
\label{tab:modality-gap-ocr-results}
\end{table*}

\begin{figure*}[t]
    \centering
    \includegraphics[width=0.85\textwidth]{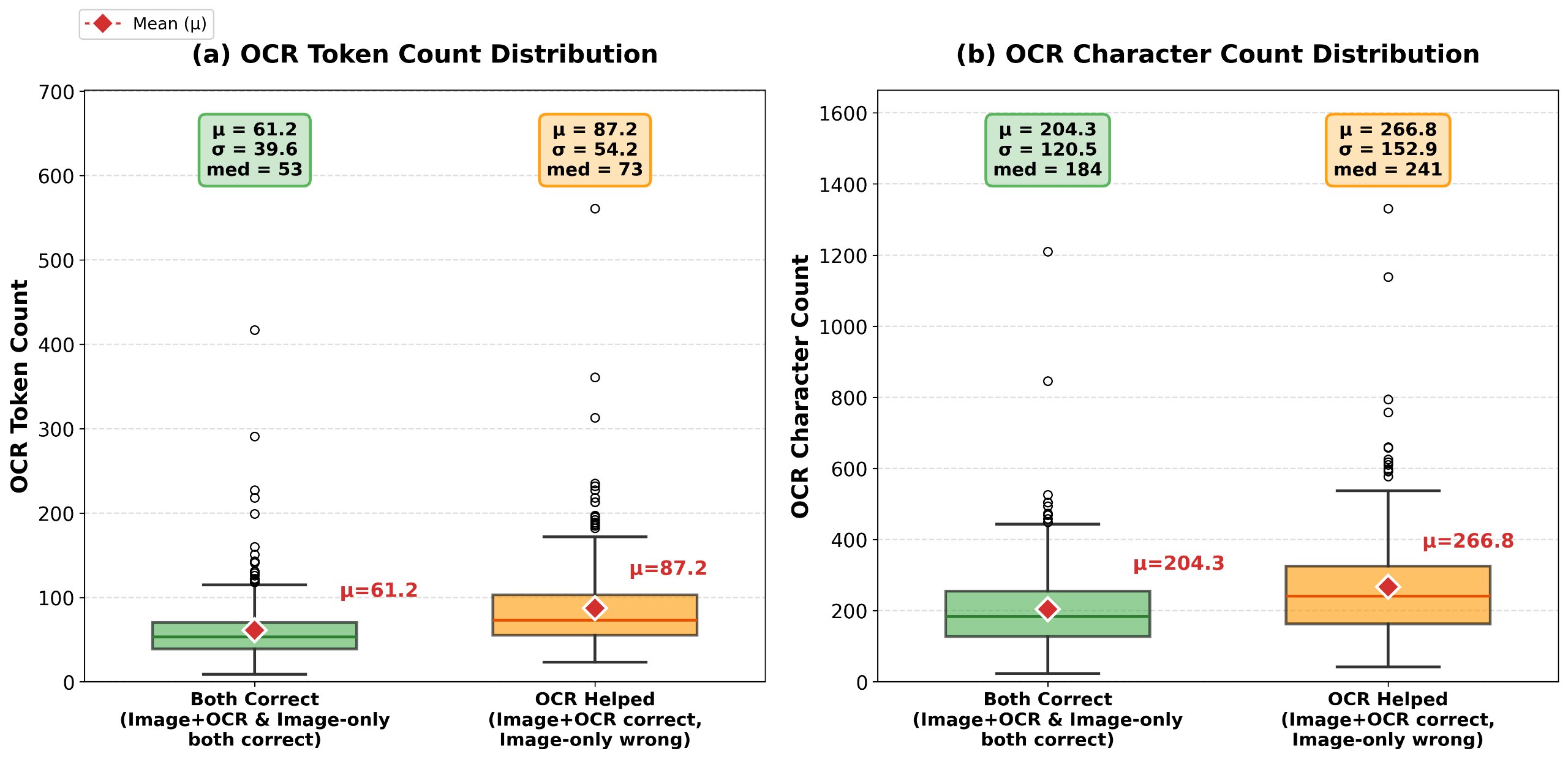}
    \caption{OCR Impact Analysis for Qwen2.5-VL-7B-Instruct (300 randomly sampled problems). Box plots show OCR text length distribution for two groups: \textit{Both Correct} (solved correctly by both image-only and image+OCR) and \textit{OCR Helped} (solved correctly \emph{only} when OCR was added). Mean token count: 61.18 (Both Correct) vs.\ 87.17 (OCR Helped); Std: 39.64 vs.\ 54.21. Mean character count: 204.31 vs.\ 266.80; Std: 120.45 vs.\ 152.85. Problems where OCR makes the difference have significantly longer text, indicating that smaller models struggle with dense, notation-heavy content from images alone.}
    \label{fig:boxplot}
\end{figure*}

Despite recent advancements, mathematical visual understanding remains a significant challenge for large vision-language models (VLMs) such as GPT-5 \citep{openai2025gpt5,lu2024mathvista,wang2024mathv,wang2025vcbench}. These tasks often involve interpreting charts, equations, and embedded text simultaneously, where even minor misinterpretations can lead to cascading errors and incorrect final answers \citep{lu2024mathvista,masry2022chartqa}. To empirically demonstrate the existence of this modality gap, we compare model performance on identical problems presented in two formats: pure text and a rendered visual format (converted via our LaTeX rendering pipeline illustrated in Figure~\ref{fig:pipeline-eval-gen}). This comparison allows us to isolate and quantify performance degradation caused by the visual modality.

We conduct this evaluation using a subset of 5{,}000 samples from our NuminaMath dataset, rendered into a visual format using our pipeline. We assess two models: Qwen2.5-VL-7B-Instruct \citep{bai2025qwen25vl} (a smaller-scale VLM) and GPT-5 \citep{openai2025gpt5} (a state-of-the-art large VLM). Table~\ref{tab:modality-gap-ocr-results} presents their performance on both formats, highlighting the modality gap across model scales. The results show that both models perform better in the text modality than in the image modality, underscoring the increased difficulty vision-language models face when interpreting and solving mathematical problems presented in visual formats. Moreover, the modality gap—the performance difference between text and image inputs—is more pronounced in smaller models such as Qwen2.5-VL-7B-Instruct than in larger models like GPT-5. This disparity may be attributed to several factors, including model scale, the sophistication of the vision encoder, and enhanced reasoning capabilities in larger architectures. Notably, tool execution primarily helps \emph{after} the model has correctly parsed mathematical content from the image; when errors are dominated by perception failures (e.g., misreading notation or layout), tool execution cannot recover from an incorrect parse, which motivates our complementary use of controlled text$\leftrightarrow$image evaluation and real-world images.

To examine whether OCR can enhance the performance of vision-language models, we apply OCR to the NuminaMath dataset using the recently released, state-of-the-art DeepSeek-OCR model \cite{wei2025deepseek}. DeepSeek-OCR comprises two key components: DeepEncoder and a DeepSeek3B-MoE-A570M decoder, and it achieves strong results across diverse OCR tasks \cite{wei2025deepseek}. We then evaluate the impact of OCR in two settings: (1) as auxiliary information alongside the original image input, and (2) as the primary input to the vision-language models. The results reveal that while both models perform strongly in the text modality, their performance in the image modality degrades, especially for Qwen2.5-VL-7B-Instruct. Interestingly, OCR-only input outperforms raw image input for both models, highlighting the importance of textual grounding \cite{shenoy2024lumos}. For Qwen, combining the image with OCR significantly boosts accuracy, whereas GPT-5 shows a decline, suggesting that OCR may introduce redundancy or noise for stronger models \cite{baek2025ocrheads}.

Our experiments highlight two strategies for reducing the text-to-image modality gap in visual mathematical reasoning: (1) teaching VLMs such as Qwen2.5-VL-7B-VisTIRA to perform tool-integrated reasoning, and (2) augmenting image inputs with OCR-based textual grounding. The results show that OCR provides substantial gains for smaller models, primarily due to their weaker visual perception, whereas larger models exhibit only marginal improvements. This suggests that OCR acts as a compensatory signal for models with limited visual capacity, while tool-integrated reasoning remains essential for scaling performance across model sizes.

We note that these OCR grounding experiments are diagnostic: as Table~\ref{tab:modality-gap-ocr-results} shows, raw OCR concatenation benefits smaller models (Qwen2.5-VL-7B: 58.77\% $\to$ 63.89\% with Image+OCR) but can be neutral or slightly harmful for stronger models (GPT-5: 73.94\% $\to$ 76.37\% with Image+OCR, yet 77.49\% with OCR-only), suggesting that the raw image signal becomes partially redundant when high-quality OCR text is available. This scale-dependent pattern motivates future work on adaptive fusion strategies such as OCR-confidence filtering and selective truncation. The paper's primary mitigation strategy remains tool-integrated supervision (VisTIRA), which provides consistent gains regardless of model scale.

To quantify when OCR grounding improves mathematical reasoning, we analyze 300 randomly sampled problems for Qwen2.5-VL-7B-Instruct across two groups: (1) \textbf{Both Correct}, problems solved correctly by both the image-only and image+OCR settings, and (2) \textbf{OCR Helped}, problems solved correctly only when OCR is added. Figure~\ref{fig:boxplot} shows box plots of OCR text length for the two groups. The results reveal that OCR-Helped problems contain significantly longer text, measured by both token and character counts. This indicates that smaller models such as Qwen2.5 struggle to fully interpret complex, detail-heavy problems from the image modality alone, leading to errors that OCR can effectively mitigate.

We further validate that the modality gap generalizes beyond Qwen2.5 by evaluating Qwen3-VL-2B and Qwen3-VL-4B on a subset of NuminaMath problems; the consistent text $>$ image trend across model families confirms the gap is a general phenomenon (see Appendix~\ref{appendix:qwen3-cross-family}, Table~\ref{tab:qwen3-modality-gap}).

\subsection{Noisy Rendered-Image Stress Test}

The rendered NuminaMath images are intentionally clean to provide a controlled lower bound on the modality gap. To evaluate robustness under more realistic visual conditions, we additionally apply a stochastic degradation pipeline to the rendered images. The pipeline applies random combinations of: Gaussian blur, Gaussian noise, brightness/contrast jitter, JPEG recompression artifacts, paper-tint coloring, and slight rotation, simulating common artifacts found in photos of printed or handwritten math. We apply this to the full 5k NuminaMath evaluation set and re-evaluate both GPT-5 and Qwen2.5-VL-7B-Instruct under identical prompting conditions.

\begin{table}[t]
\centering
\renewcommand{\arraystretch}{1.15}
\resizebox{\columnwidth}{!}{%
\begin{tabular}{lccc}
\hline
\textbf{Model} & \textbf{Text (\%)} & \textbf{Image Clean (\%)} & \textbf{Image Noisy (\%)} \\
\hline
Qwen2.5-VL-7B-Instruct & 64.19 & 58.77 & 59.26 \\
GPT-5 & 79.94 & 73.94 & 72.65 \\
\hline
\end{tabular}}
\vspace{0.5em}
\caption{Noisy rendered-image stress test on NuminaMath (5k). GPT-5 shows a clear drop under noise (73.94 $\to$ 72.65), widening its modality gap from +6.00 to +7.29. Qwen2.5-VL-7B remains stable (58.77 $\to$ 59.26). A substantial modality gap persists for both models regardless of image quality.}
\label{tab:noisy-stress-test}
\end{table}

GPT-5's accuracy drops under noise (73.94\% $\to$ 72.65\%), widening its modality gap from +6.00 to +7.29 points and confirming that the clean-render setting provides a conservative lower bound. Qwen2.5-VL-7B-Instruct is essentially unaffected (58.77\% $\to$ 59.26\%), which we attribute to a visual-parsing ceiling: its accuracy is already constrained by layout and formula recognition, so pixel-level noise adds no further errors. In both cases, a substantial modality gap persists, demonstrating that the gap is structural rather than an artifact of image quality (see Figure~\ref{fig:noisy-examples} for example noisy images). To support community evaluation under realistic conditions, we release the full noisy rendered NuminaMath dataset (360k images) alongside the clean 360k renders.

\section{Limitations}

This work has a few limitations. First, while our LaTeX-based rendering pipeline enables controlled modality gap analysis, synthetically rendered images do not fully capture the noise, distortions, and stylistic diversity of real-world handwritten or photographed mathematical content, which may limit generalization. Second, our tool-integrated trajectories are generated using strong teacher models and automatically filtered, which may introduce bias toward specific reasoning styles and reduce diversity in solution strategies. Finally, although OCR grounding improves performance for smaller VLMs, its effectiveness diminishes for larger models and can introduce redundant or noisy inputs, suggesting that more adaptive grounding mechanisms are needed. Addressing these limitations will be important for scaling visual mathematical reasoning to broader, real-world settings.

\bibliography{references}


\clearpage
\onecolumn

\appendix

\renewcommand{\thefigure}{A.\arabic{figure}}
\setcounter{figure}{0}
\renewcommand{\thetable}{A.\arabic{table}}
\setcounter{table}{0}

\section{Appendix}
\label{appendix:Appendix}

\subsection{Supporting Figures}
\label{appendix:supporting-figures}

\vspace{-0.5em}

\begin{figure}[H]
\centering
\begin{tikzpicture}[font=\large, scale=0.85, every node/.style={scale=0.85}]
\node[pillBlue, minimum width=4.0cm] (problem) {Math Problem Image};
\node[panel=7cm, right=\distInputLLM of problem.center] (llm) {};
\coordinate (C) at ($(llm.north west)!0.5!(llm.south east)$);
\def\r{1.40cm}
\node[pillPink] (rationale) at ($(C)+(-\r,1)$) {Rationale ($\rho_i$)};
\node[pillGreen] (program)  at ($(C)+(\r+\gapRP,0.5)$) {Program ($\mathcal{A}_i$)};
\node[pillBlue] (output)    at ($(C)+(-1,-1.0cm)$) {Output ($\mathcal{O}_i$)};
\draw[cycleArrow, bend left=18] (rationale) to (program);
\draw[cycleArrow, bend left=18] (program)   to (output);
\draw[cycleArrow, bend left=18] (output)    to (rationale);
\node[font=\Large, below=0.2cm of llm] {using a VLM};
\node[panel=3.8cm, anchor=west] (finetune) at ($(llm.east)+(\distLLMFT,0)$) {};
\node[font=\normalsize, anchor=north] (td_label) at ($(finetune.north)+(0,-0.2)$) {\textbf{Training Data}};
\node[pillBlue, scale=0.65, minimum width=2cm, anchor=north] (td_img) 
    at ($(finetune.center)+(0, 0.3)$) {Image};
\node[draw=black!70, fill=white, rounded corners=3pt, scale=0.65, minimum width=2cm, minimum height=1.5em, anchor=north] (td_traj) 
    at ($(td_img.south)+(0,-0.4)$) {Trajectory ($\rho_i, \mathcal{A}_i$)};
\node[font=\footnotesize, text=gray] at ($(td_img.south)!0.5!(td_traj.north)$) {+};
\draw[arrow] (problem) -- (llm);
\draw[trajArrow] 
  ($(llm.east)+(\arrowPad,0)$) -- 
  node[font=\normalsize, above=3pt, align=center] {Valid Trajectories\\ \small (Filtering)} 
  ($(finetune.west)+(-\arrowPad,0)$);
\end{tikzpicture}
\vspace{-0.5em}
\caption{This figure illustrates the VisTIRA data generation pipeline. We begin by prompting powerful vision language models (e.g., GPT-5, Gemini) to produce tool-integrated reasoning trajectories, which we refer to as VisTIRA data. This synthetic dataset is then used to fine-tune a smaller VLM, enabling it to function as a mathematical reasoning agent.}
\label{fig:diagram-VisTIRA}
\end{figure}

\vspace{-1em}

\begin{figure}[H]
\centering
\begin{tikzpicture}[
  font=\small,
  node distance=10mm and 12mm,
  >=Stealth,
  every node/.style={align=center},
  data/.style={draw, rounded corners=2pt, fill=yellow!20, inner sep=5pt},
  proc/.style={draw, rounded corners=2pt, fill=orange!20, inner sep=5pt},
  art/.style ={draw, rounded corners=2pt, fill=blue!10,   inner sep=5pt},
  groupbox/.style={draw, rounded corners=2pt, inner sep=8pt},
  solidlink/.style={->, thick},
  dashedlink/.style={->, dashed, thick},
]

\node[data] (textmath) {NuminaMath Dataset\\ Problem (text), Answer (Text)};
\node[groupbox, fit=(textmath), label=above:{\footnotesize Data Source}] (datasources) {};

\node[proc, right=22mm of textmath] (llm) {LLM};
\node[proc, above=6mm of llm] (prompt) {Modality conversion prompt\\(structure-preserving rules)};
\node[art,  right=8mm of llm] (latexcode) {\LaTeX{} code\\(compilable)};
\node[proc, right=8mm of latexcode] (latexeng) {\LaTeX{} engine};

\node[groupbox, fit=(prompt)(llm)(latexcode),
      label=above:{\footnotesize Modality Conversion via LLM (\ref{appendix:text-to-latex})}] (llmgroup) {};

\node[data, below=10mm of latexeng] (eval_img) {Problem \\ (Image)};
\node[data, right=10mm of eval_img] (eval_ans) {Answer \\ (Text)};

\node[groupbox, fit=(eval_img)(eval_ans), inner sep=10pt, label=below:{\footnotesize Evaluation Dataset}] (finalgroup) {};

\draw[solidlink] (textmath) -- (llm);
\draw[solidlink] (prompt) -- (llm);
\draw[solidlink] (llm) -- (latexcode);
\draw[solidlink] (latexcode) -- (latexeng);
\draw[solidlink] (latexeng) -- node[right, font=\footnotesize] {Render} (eval_img);

\draw[solidlink] (textmath.south) -- ++(0,-2.1cm) -| 
    (eval_ans.south) node[above, pos=0.25, font=\footnotesize] {Preserve Ground Truth Answer};

\end{tikzpicture}
\vspace{-0.5em}
\caption{Data generation pipeline for creating the image-modality evaluation dataset from NuminaMath. Text-only math problems are processed by an LLM to generate compilable \LaTeX{} code, which is rendered into typeset problem images. Crucially, the original ground-truth answers from NuminaMath are preserved and paired directly with the generated images to form the final evaluation set, bypassing the reasoning generation step.}
\label{fig:pipeline-eval-gen}
\end{figure}

\vspace{-1em}

\begin{figure}[H]
   \centering
    \begin{subfigure}{0.45\textwidth}
       \centering
       \includegraphics[width=\linewidth]{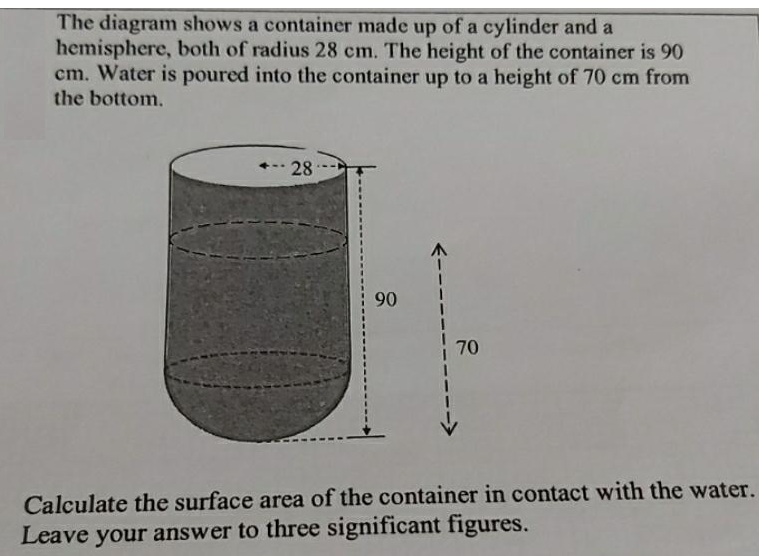}
       \caption{Real-world example math problem}
    \end{subfigure}
    \hfill
    \begin{subfigure}{0.45\textwidth}
       \centering
       \includegraphics[width=\linewidth]{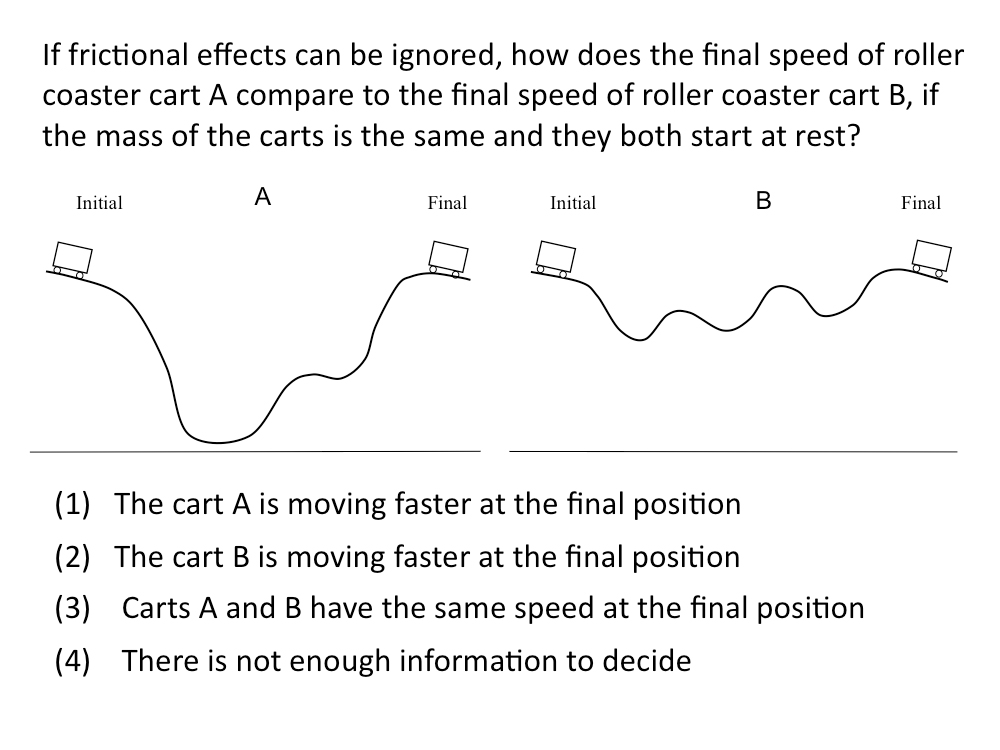}
       \caption{Real world example physics problem}
   \end{subfigure}
    
    \caption{Examples of real-world challenging visual question answering tasks: 
   (Left) a math homework problem requiring symbolic reasoning; 
    (Right) a physics homework problem requiring interpretation of both text and diagram.}
   \label{fig:hm_problems}
\end{figure}

\subsection{VisTIRA Prompt for Data Generation}
\label{appendix:VisTIRA}

The following prompt is used to generate tool-integrated reasoning trajectories from vision-language models:

\begin{tcolorbox}[colback=gray!5, colframe=gray!80!black, title=VisTIRA Data Generation Prompt, breakable]
\small
\begin{verbatim}
Integrate step-by-step reasoning and Python code to solve problems from images 
using the following guidelines:

# Steps

1. Look at the provided image to understand the problem.
2. Analyze the question from the image.
3. Write a Python function to solve the problem; the function should not take 
   any arguments.
4. Present the final result in LaTeX using a '\boxed{}' without any units.
5. Utilize the 'pi' symbol and 'Rational' from Sympy for $\pi$ and fractions, 
   and simplify all fractions and square roots without converting them to 
   decimal values.
6. Ensure the reasoning steps are clear and logical.
7. Only solve the first question if multiple questions are present in the image.
8. Return `<no_response>` if the question cannot be solved for any reason.
9. After each Python code block, include an output block showing the function's 
   result exactly like this:
   ```python
   [your python code here]
   ```
   ```output
   [the exact output from running the function]
   ```
10. IMPORTANT: Write the Python code solution in a code block with these 
    requirements:
    - Always store the function's return value in a variable
    - ALWAYS use print() to show the output, never leave just the variable name
11. Only generate Python code for mathematical problems. For other subjects 
    provide a direct answer unless Python programming is specifically requested.
12. ***VERY IMPORTANT***: You must continue the solution from where it was 
    left off, completing the remaining part without altering or repeating 
    any previously written sections.

# Output Format

- The final result should be presented in LaTeX using a '\boxed{}' without 
  any units.
- Include Python code blocks and output blocks as specified.
\end{verbatim}
\end{tcolorbox}

\vspace{1em}
\noindent\textbf{Example 1: Spherical Coordinates Conversion}

\begin{tcolorbox}[colback=blue!5, colframe=blue!50!black, breakable]
\small
\textbf{Question:} Convert the point $(0, -3\sqrt{3}, 3)$ in rectangular coordinates to spherical coordinates. Enter your answer in the form $(\rho,\theta,\phi)$, where $\rho > 0$, $0 \le \theta < 2\pi$, and $0 \le \phi \le \pi$.

\textbf{Solution:}

To convert $(x, y, z)$ to spherical coordinates $(\rho, \theta, \phi)$:
$\rho = \sqrt{x^2 + y^2 + z^2}$, $\theta = \arctan\frac{y}{x}$, $\phi = \arccos\frac{z}{\rho}$

\begin{verbatim}
from sympy import sqrt, atan2, acos, pi

def rectangular_to_spherical():
    x, y, z = 0, -3*sqrt(3), 3
    rho = sqrt(x**2 + y**2 + z**2)
    theta = atan2(y, x)
    phi = acos(z/rho)
    return rho, theta, phi

print(rectangular_to_spherical())
\end{verbatim}

\textbf{Output:} \texttt{(6, -pi/2, pi/3)}

Since $\theta = -\pi/2$ doesn't meet $0 \le \theta < 2\pi$, we add $2\pi$:

\textbf{Final Answer:} $\boxed{\left(6, \frac{3\pi}{2}, \frac{\pi}{3}\right)}$
\end{tcolorbox}

\vspace{1em}
\noindent\textbf{Example 2: Binary Arithmetic}

\begin{tcolorbox}[colback=blue!5, colframe=blue!50!black, breakable]
\small
\textbf{Question:} Compute $1011_2 + 101_2 - 1100_2 + 1101_2$. Express your answer in base 2.

\begin{verbatim}
def binary_sum_diff():
    num1 = int("1011", 2)
    num2 = int("101", 2)
    num3 = int("1100", 2)
    num4 = int("1101", 2)
    result = num1 + num2 - num3 + num4
    return format(result, "b")

print(binary_sum_diff())
\end{verbatim}

\textbf{Output:} \texttt{10001}

\textbf{Final Answer:} $\boxed{10001_2}$
\end{tcolorbox}

\vspace{1em}
\noindent\textbf{Example 3: Solving Inequalities}

\begin{tcolorbox}[colback=blue!5, colframe=blue!50!black, breakable]
\small
\textbf{Question:} Solve the inequality $\dfrac{x^2 - 25}{x + 5} < 0$.

\begin{verbatim}
from sympy import symbols, simplify
from sympy.solvers.inequalities import solve_univariate_inequality
from sympy.core.relational import StrictLessThan

def solve_inequality():
    x = symbols('x')
    expression = (x**2 - 25) / (x + 5)
    inequality = StrictLessThan(expression, 0)
    solution = solve_univariate_inequality(inequality, x, relational=False)
    return simplify(solution)

print(solve_inequality())
\end{verbatim}

\textbf{Output:} \texttt{Union(Interval.open(-oo, -5), Interval.open(-5, 5))}

\textbf{Final Answer:} $\boxed{(-\infty,-5)\cup(-5,5)}$
\end{tcolorbox}

\subsection{VisTIRA Trajectory Examples}
\label{appendix:VisTIRA-examples}

This section presents sample VisTIRA trajectories generated by GPT-4o, demonstrating the tool-integrated reasoning approach.

\vspace{1em}
\noindent\textbf{VisTIRA Trajectory Example 1}
\label{appendix:VisTIRA-example1}

\begin{tcolorbox}[colback=gray!5, colframe=gray!80!black, title=VisTIRA Trajectory Example 1, breakable]
\raggedright\small
\begin{center}
\includegraphics[width=0.5\textwidth]{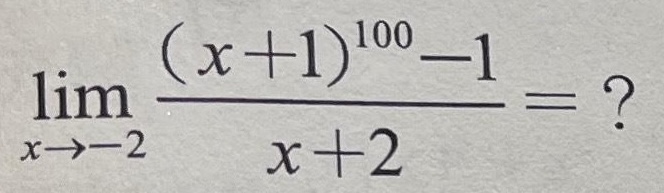}\\[6pt]
\textbf{Example Image:} Limit problem involving L'H\^{o}pital's Rule.
\end{center}

\vspace{6pt}
\textbf{VisTIRA Trajectory:} To solve the given limit problem, we need to evaluate:
\[
\lim_{{x \to -2}} \frac{(x+1)^{100} - 1}{x+2}
\]

\textbf{Steps:}
\begin{enumerate}
    \item \textbf{Direct substitution}: $\frac{(-1)^{100} - 1}{0} = \frac{1 - 1}{0} = \frac{0}{0}$ --- This is an indeterminate form.
    \item \textbf{Apply L'H\^{o}pital's Rule}: Since the form is $\frac{0}{0}$, differentiate numerator and denominator.
    \item \textbf{Differentiate}: Numerator: $100(x+1)^{99}$; Denominator: $1$
    \item \textbf{Re-evaluate the limit}: $\lim_{{x \to -2}} \frac{100(x+1)^{99}}{1} = 100(-1)^{99} = -100$
\end{enumerate}

\textbf{Python code:}
\begin{verbatim}
from sympy import symbols, limit, diff

def solve_limit():
    x = symbols('x')
    numerator = (x + 1)**100 - 1
    denominator = x + 2
    num_diff = diff(numerator, x)
    den_diff = diff(denominator, x)
    result = limit(
        num_diff / den_diff, x, -2)
    return result

result = solve_limit()
print(result)
\end{verbatim}

\textbf{Output:} \texttt{-100}

\textbf{Final answer:} $\boxed{-100}$
\end{tcolorbox}

\vspace{1em}
\noindent\textbf{VisTIRA Trajectory Example 2}
\label{appendix:VisTIRA-example2}

\begin{tcolorbox}[colback=gray!5, colframe=gray!80!black, title=VisTIRA Trajectory Example 2, breakable]
\begin{center}
\includegraphics[width=0.5\textwidth]{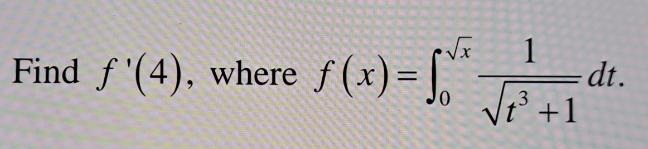}\\[6pt]
\textbf{Example Image:} Integral-based derivative problem using the Fundamental Theorem of Calculus.
\end{center}

\vspace{6pt}
\textbf{VisTIRA Trajectory:} To solve the problem in the image, we need to compute:
\[
f'(4) \quad \text{where} \quad f(x) = \int_0^{\sqrt{x}} \frac{1}{\sqrt{t^3 + 1}} \, dt
\]

\textbf{Steps:}
\begin{enumerate}
    \item \textbf{Apply the Fundamental Theorem of Calculus and Chain Rule}: $f'(x) = \frac{1}{\sqrt{(\sqrt{x})^3 + 1}} \cdot \frac{d}{dx}(\sqrt{x})$
    \item \textbf{Simplify the expression}: $f'(x) = \frac{1}{\sqrt{x^{3/2} + 1}} \cdot \frac{1}{2\sqrt{x}} = \frac{1}{2\sqrt{x} \sqrt{x^{3/2} + 1}}$
    \item \textbf{Evaluate at $x = 4$}: $f'(4) = \frac{1}{2 \cdot \sqrt{4} \cdot \sqrt{4^{3/2} + 1}} = \frac{1}{4 \cdot \sqrt{9}} = \frac{1}{4 \cdot 3} = \frac{1}{12}$
\end{enumerate}

\textbf{Python code:}
\begin{verbatim}
from sympy import symbols, sqrt, diff

def find_derivative_at_4():
    x = symbols('x')
    t = symbols('t')
    integrand = 1 / sqrt(t**3 + 1)
    upper_limit = sqrt(x)
    integrand_at_upper_limit = integrand.subs(t, upper_limit)
    derivative = diff(integrand_at_upper_limit, x)
    derivative_at_4 = derivative.subs(x, 4)
    return derivative_at_4.simplify()

result = find_derivative_at_4()
print(result)
\end{verbatim}

\textbf{Output:} \texttt{1/12}

\textbf{Final answer:} $\boxed{\frac{1}{12}}$
\end{tcolorbox}

\subsection{Prompt for Converting Text-to-Text Math Problems to Image-to-Text Problems}
\label{appendix:text-to-latex}

\begin{tcolorbox}[colback=gray!5, colframe=gray!80!black, title=Text-to-LaTeX Conversion Prompt, breakable]
\small
\begin{verbatim}
I will provide you with a text string that may contain LaTeX equations and text.
Your task is to analyze it, rewrite it as correctly executable LaTeX code, and 
include all necessary packages to ensure it runs without errors. Do not alter 
the original content; only make adjustments required for compatibility with a 
LaTeX engine.

IMPORTANT: You must return the response as a Python string without any prefix 
or suffix.

EXAMPLE:
---
Text:
Given that the function f(x)=sin($\pi$-\omega x)cos \omega x + cos^2\omega x (\omega > 0) 
has a minimum positive period of $\pi$.
(I) Find the value of $\omega$;
(II) Find the minimum value of y=g(x) on the interval [0, $\pi$/16].

ANSWER:
\documentclass[12pt]{article}
\usepackage{amsmath}
\usepackage{amssymb}
\usepackage{geometry}
\geometry{a4paper, margin=1in}
\begin{document}

Given that the function 
\[
f(x) = \sin (\pi - \omega x)\cos \omega x + \cos^{2}\omega x \quad (\omega > 0)
\]
has a minimum positive period of $\pi$.

\begin{enumerate}
    \item[(I)] Find the value of $\omega$;
    \item[(II)] Find the minimum value of $y = g(x)$ on $\left[0, \frac{\pi}{16}\right]$.
\end{enumerate}

\end{document}
\end{verbatim}
\end{tcolorbox}

\subsection{Text-to-Image Converted Examples}
\label{appendix:text_to_image_examples}

This section presents examples of mathematical problems converted from text modality to image modality using our LaTeX rendering pipeline.

\vspace{1em}

\begin{tcolorbox}[title=Problem and LaTeX Source Example 1, colback=gray!5, colframe=gray!80!black, fonttitle=\bfseries, breakable]
\small
\begin{center}
\includegraphics[width=0.5\textwidth]{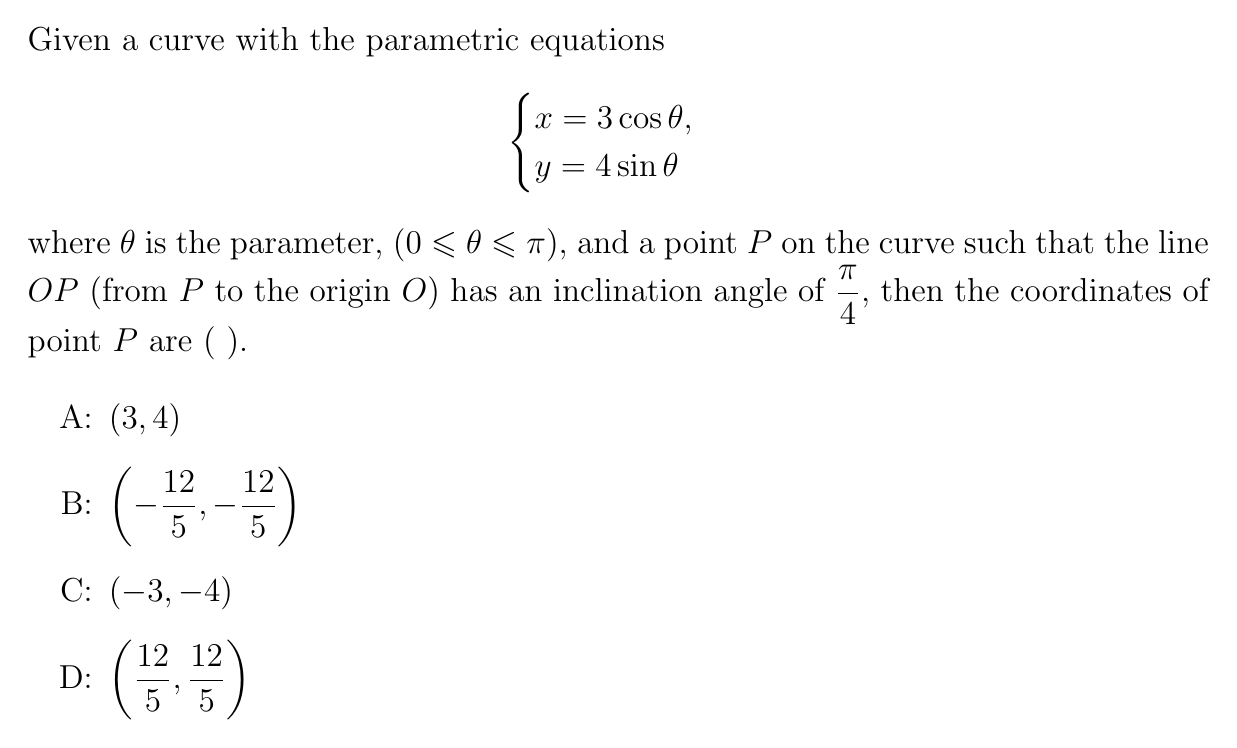}\\[6pt]
\textbf{Example Image 1:} LaTeX-rendered version of the problem statement in image modality.
\end{center}

\vspace{8pt}
\noindent
\textbf{Problem in Text Modality (shown as raw text):}
\begin{verbatim}
Given a curve with the parametric equations
\[
\begin{cases} 
x=3\cos \theta, \\
y=4\sin \theta 
\end{cases}
\]
where $\theta$ is the parameter, $(0\leqslant \theta\leqslant \pi)$, 
and a point $P$ on the curve such that the line $OP$ (from $P$ to the origin $O$) 
has an inclination angle of $\dfrac{\pi}{4}$, then the coordinates of point $P$
are ( ).

A: $(3,4)$
B: $\left( -\dfrac{12}{5},- \dfrac{12}{5} \right)$
C: $(-3,-4)$
D: $\left( \dfrac{12}{5}, \dfrac{12}{5} \right)$
\end{verbatim}

\vspace{8pt}
\noindent
\textbf{LaTeX Code Used for Rendering:}
\begin{verbatim}
\documentclass[12pt]{article}
\usepackage{amsmath}
\usepackage{amssymb}
\usepackage{amsfonts}
\usepackage{geometry}
\geometry{a4paper, margin=1in}
\begin{document}

Given a curve with the parametric equations
\[
\begin{cases} 
x = 3\cos \theta, \\
y = 4\sin \theta 
\end{cases}
\]
where $\theta$ is the parameter, $(0 \leqslant \theta \leqslant \pi)$, 
and a point $P$ on the curve such that the line $OP$ (from $P$ to the origin $O$) 
has an inclination angle of $\dfrac{\pi}{4}$, then the coordinates of
point $P$ are ( ).

\begin{itemize}
    \item[A:] $(3,4)$
    \item[B:] $\left( -\dfrac{12}{5}, -\dfrac{12}{5} \right)$
    \item[C:] $(-3,-4)$
    \item[D:] $\left( \dfrac{12}{5}, \dfrac{12}{5} \right)$
\end{itemize}

\end{document}
\end{verbatim}
\end{tcolorbox}

\vspace{1em}

\begin{tcolorbox}[title=Problem and LaTeX Source Example 2, colback=gray!5, colframe=gray!80!black, fonttitle=\bfseries, breakable]
\small
\begin{center}
\includegraphics[width=0.7\textwidth]{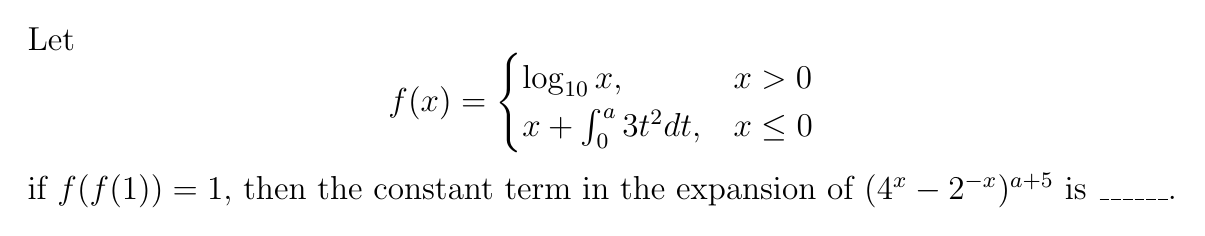}\\[6pt]
\textbf{Example Image 2:} LaTeX-rendered version of the problem statement in image modality.
\end{center}

\vspace{8pt}
\noindent
\textbf{Problem in Text Modality (shown as raw text):}
\begin{verbatim}
Let $f(x) = \begin{cases} \log_{10}x, & x>0 \\ x+ \int_{0}^{a}3t^{2}dt, & x\leq 0
\end{cases}$, 
if $f(f(1))=1$, then the constant term in the expansion of
$(4^x-2^{-x})^{a+5}$ is _____.
\end{verbatim}

\vspace{8pt}
\noindent
\textbf{LaTeX Code Used for Rendering:}
\begin{verbatim}
\documentclass[12pt]{article}
\usepackage{amsmath}
\usepackage{amssymb}
\usepackage{amsfonts}
\usepackage{geometry}
\geometry{a4paper, margin=1in}
\begin{document}

Let 
\[
f(x) = 
\begin{cases} 
\log_{10}x, & x>0 \\ 
x+ \int_{0}^{a}3t^{2}dt, & x\leq 0 
\end{cases}
\]
if $f(f(1))=1$, then the constant term in the expansion of $(4^x-2^{-x})^{a+5}$ is
_____.

\end{document}
\end{verbatim}
\end{tcolorbox}

\vspace{1em}

\begin{tcolorbox}[title=Problem and LaTeX Source Example 3, colback=gray!5, colframe=gray!80!black, fonttitle=\bfseries, breakable]
\small
\begin{center}
\includegraphics[width=0.5\textwidth]{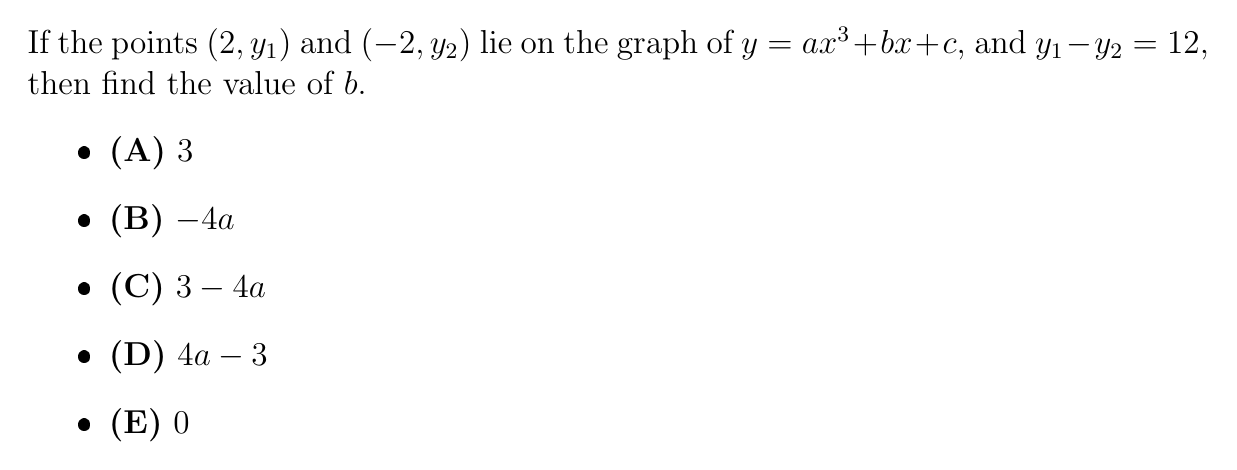}\\[6pt]
\textbf{Example Image 3:} LaTeX-rendered version of the problem statement in image modality.
\end{center}

\vspace{8pt}
\noindent
\textbf{Problem in Text Modality (shown as raw text):}
\begin{verbatim}
If the points $(2,y_1)$ and $(-2,y_2)$ lie on the graph of $y=ax^3+bx+c$, 
and $y_1-y_2=12$, then find the value of $b$.
- **(A)** \(3\)
- **(B)** \( -4a\)
- **(C)** \(3 - 4a\)
- **(D)** \(4a - 3\)
- **(E)** \(0\)
\end{verbatim}

\vspace{8pt}
\noindent
\textbf{LaTeX Code Used for Rendering:}
\begin{verbatim}
\documentclass[12pt]{article}
\usepackage{amsmath}
\usepackage{amssymb}
\usepackage{amsfonts}
\usepackage{geometry}
\geometry{a4paper, margin=1in}
\begin{document}

If the points $(2,y_1)$ and $(-2,y_2)$ lie on the graph of $y = ax^3 + bx + c$, 
and $y_1 - y_2 = 12$, then find the value of $b$.
\begin{itemize}
    \item \textbf{(A)} \(3\)
    \item \textbf{(B)} \( -4a\)
    \item \textbf{(C)} \(3 - 4a\)
    \item \textbf{(D)} \(4a - 3\)
    \item \textbf{(E)} \(0\)
\end{itemize}

\end{document}
\end{verbatim}
\end{tcolorbox}

\vspace{1em}

\begin{tcolorbox}[title=Problem and LaTeX Source Example 4, colback=gray!5, colframe=gray!80!black, fonttitle=\bfseries, breakable]
\small
\begin{center}
\includegraphics[width=0.7\textwidth]{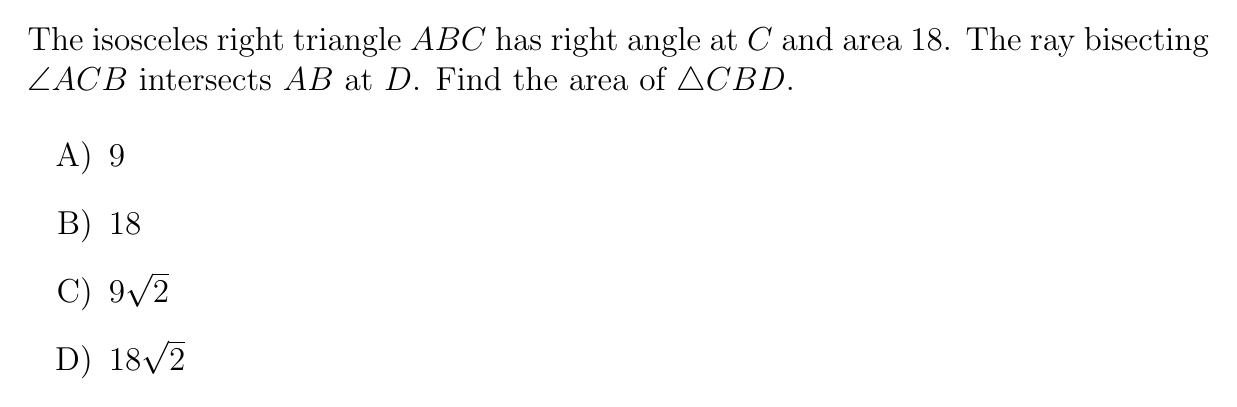}\\[6pt]
\textbf{Example Image 4:} LaTeX-rendered version of the problem statement in image modality.
\end{center}

\vspace{8pt}
\noindent
\textbf{Problem in Text Modality (shown as raw text):}
\begin{verbatim}
The isosceles right triangle \(ABC\) has right angle at \(C\) and area \(18\). 
The ray bisecting \(\angle ACB\) intersects \(AB\) at \(D\). 
Find the area of \(\bigtriangleup CBD\).  
Options:  
A) \(9\)  
B) \(18\)  
C) \(9\sqrt{2}\)  
D) \(18\sqrt{2}\)
\end{verbatim}

\vspace{8pt}
\noindent
\textbf{LaTeX Code Used for Rendering:}
\begin{verbatim}
\documentclass[12pt]{article}
\usepackage{amsmath}
\usepackage{amssymb}
\usepackage{amsfonts}
\usepackage{geometry}
\geometry{a4paper, margin=1in}
\begin{document}

The isosceles right triangle $ABC$ has right angle at $C$ and area $18$. 
The ray bisecting $\angle ACB$ intersects $AB$ at $D$. 
Find the area of $\bigtriangleup CBD$.

\begin{enumerate}
    \item[A)] $9$
    \item[B)] $18$
    \item[C)] $9\sqrt{2}$
    \item[D)] $18\sqrt{2}$
\end{enumerate}

\end{document}
\end{verbatim}
\end{tcolorbox}

\subsection{Dataset Statistics \& Trajectory Properties}
\label{appendix:dataset-stats}

\begin{tcolorbox}[colback=gray!5, colframe=gray!80!black, breakable, sharp corners, boxrule=0.8pt, left=4pt, right=4pt, top=4pt, bottom=4pt]
\textbf{SnapAsk VisTIRA Trajectory Statistics.}
We generated VisTIRA trajectories for approximately 303k SnapAsk images using
GPT-4o as the teacher model (Algorithm~\ref{alg:tool_inference}, $n=2$, greedy
decoding). After filtering via CoT$\leftrightarrow$tool-execution answer
verification, 147{,}948 trajectories were retained for training
(verification rate $\approx$48.9\%). Key properties of the retained
trajectories:\vspace{0.3em}
\begin{itemize}
  \item \textbf{Assistant response length (Qwen2.5 tokenizer):} mean 523.4 tokens; median 489; P90 814; P95 929.
  \item \textbf{Tool-use structure:} 1 Python code block per trajectory.
  \item \textbf{Explicit numbered reasoning steps:} mean 3.17; median 3; P90 6; P95 7.
  \item \textbf{Code library usage (import-based, over all trajectories):} SymPy 85.9\%; NumPy 0.09\%; \texttt{math} 0.82\%.
\end{itemize}\vspace{0.3em}
\textbf{Filtering criteria.} A trajectory is \emph{accepted} if and only if
the final \texttt{\textbackslash boxed\{\}} answer produced by tool execution
exactly matches the answer obtained by the model's own chain-of-thought
(CoT) reasoning on the same image. This dual-path verification ensures
internal consistency and filters out trajectories where the code silently
produces an incorrect result or where the CoT rationale diverges from the
computational path.
\end{tcolorbox}

\vspace{0.5em}

\begin{tcolorbox}[colback=gray!5, colframe=gray!80!black, breakable, sharp corners, boxrule=0.8pt, left=4pt, right=4pt, top=4pt, bottom=4pt]
\textbf{NuminaMath Rendered Image Statistics.}
From the full NuminaMath-CoT corpus ($\sim$860k problems), we converted
problems into rendered images via our LaTeX pipeline. Key dataset-level
statistics reported in the paper:\vspace{0.3em}
\begin{itemize}
  \item \textbf{Source corpus (text-only):} $\sim$860k problems.
  \item \textbf{Rendered release (clean):} 360k images.
  \item \textbf{Rendered release (noisy):} 360k images, generated via a stochastic degradation pipeline (Gaussian blur, Gaussian noise, brightness/contrast jitter, JPEG recompression artifacts, paper-tint coloring, slight rotation).
  \item \textbf{Evaluation subset:} 5{,}000 rendered problems used in the main modality-gap and noisy stress-test experiments (Table~\ref{tab:noisy-stress-test}).
  \item \textbf{Stress-test setting:} we use Severity~2 (Medium) degradations for the noisy evaluation reported in Table~\ref{tab:noisy-stress-test}.
\end{itemize}
\end{tcolorbox}

\begin{tcolorbox}[colback=gray!5, colframe=gray!80!black, breakable, boxrule=0.8pt]
\begin{figure}[H]
\centering
\includegraphics[width=0.75\textwidth]{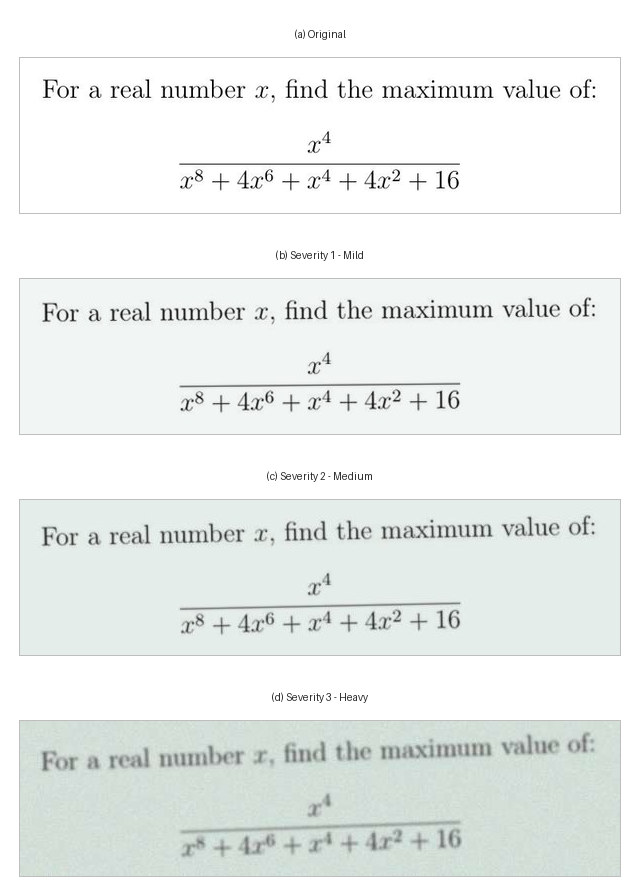}
\caption{Illustration of our stochastic image degradation pipeline applied to a rendered NuminaMath problem at increasing severity levels. (a)~Original clean render produced by our LaTeX pipeline. (b)~Severity~1 (Mild): subtle noise and slight tonal shift. (c)~Severity~2 (Medium): visible paper tint, light blur, and reduced contrast. (d)~Severity~3 (Heavy): strong background discoloration, pronounced noise, and degraded text legibility. In our stress-test evaluation we use Severity~2 (Medium) degradations to approximate realistic scan and photo conditions while keeping the problem text readable.}
\label{fig:noisy-examples}
\end{figure}
\end{tcolorbox}

\subsection{Cross-Family Modality Gap Validation}
\label{appendix:qwen3-cross-family}

To validate that the modality gap generalizes beyond Qwen2.5, we evaluate two additional models from a different generation---Qwen3-VL-2B and Qwen3-VL-4B---on a random subset of 814 NuminaMath problems. Table~\ref{tab:qwen3-modality-gap} shows the results. While the gap is notably smaller for Qwen3-VL models (1--2\%) than for Qwen2.5-VL-7B (5.4\%), the consistent text $>$ image trend across different model families, architectures, and scales confirms that the modality gap is a general phenomenon, not an artifact of a single model.

\begin{table}[H]
\centering
\renewcommand{\arraystretch}{1.15}
\begin{tabular}{lccc}
\hline
\textbf{Model} & \textbf{Text (\%)} & \textbf{Image (\%)} & \textbf{Gap} \\
\hline
Qwen3-VL-2B (814 subset) & 86.46 & 84.26 & +2.20 \\
Qwen3-VL-4B (814 subset) & 90.66 & 89.59 & +1.07 \\
\hline
Qwen2.5-VL-7B-Instruct (5k) & 64.19 & 58.77 & +5.42 \\
GPT-5 (5k) & 79.94 & 73.94 & +6.00 \\
\hline
\end{tabular}
\vspace{0.5em}
\caption{Modality gap across model families and scales on NuminaMath. Qwen3-VL models are evaluated on a random 814-problem subset; Qwen2.5 and GPT-5 on the full 5k set. The gap persists across all models, though it is smaller for Qwen3-VL, which we attribute to stronger vision encoders and training advances.}
\label{tab:qwen3-modality-gap}
\end{table}

\subsection{Qwen2.5-VL-7B-VisTIRA Example Inference}
\label{app:vistira}

This section presents comparison examples between the base Qwen2.5-VL-7B-Instruct model and the fine-tuned Qwen2.5-VL-7B-VisTIRA model.

\vspace{1em}
\noindent\textbf{Example 1: Modular Arithmetic Sequence}

\begin{tcolorbox}[colback=green!5, colframe=green!50!black, title=Base Model Fails -- Fine-Tuned Model Succeeds, fonttitle=\bfseries, breakable]
\small
\begin{center}
\includegraphics[width=0.6\textwidth]{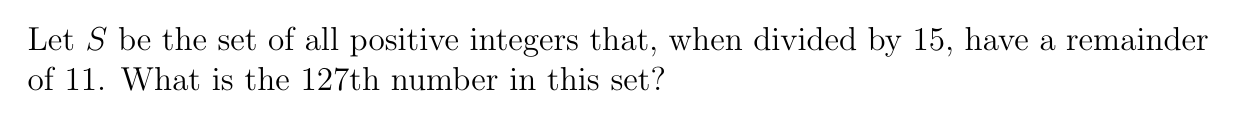}\\[6pt]
\textbf{Problem:} Modular arithmetic sequence problem where the base model fails and the fine-tuned model succeeds.
\end{center}

\vspace{8pt}
\noindent
\textbf{Base Model Output:} \textcolor{red}{$\times$}

Find the 127th number in the sequence:
\[
n = 15k + 11
\]
Uses:
\[
k = 126 \Rightarrow n = 15 \cdot 126 + 11 = 1890 + 11 = \boxed{1891}
\]

\vspace{1em}
\textbf{Fine-Tuned Model Output:} \textcolor{green}{$\checkmark$}

Same formula:
\[
n = 15k + 11,\quad k = 126
\]

Python verification:
\begin{verbatim}
def find_127th_number():
    k = 126
    n = 15 * k + 11
    return n

print(find_127th_number())
\end{verbatim}

Output: \texttt{1901}

Final result: $\boxed{1901}$

\vspace{1em}
\textbf{Ground Truth:} $\boxed{1901}$
\end{tcolorbox}

\vspace{1em}
\noindent\textbf{Example 2: Divisibility Problem}

\begin{tcolorbox}[colback=green!5, colframe=green!50!black, title=Base Model Fails -- Fine-Tuned Model Succeeds, fonttitle=\bfseries, breakable]
\small
\begin{center}
\includegraphics[width=0.55\textwidth]{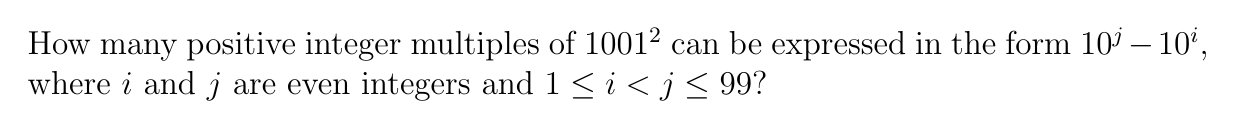}\\[6pt]
\textbf{Problem:} Divisibility problem where the base model fails and the fine-tuned model succeeds.
\end{center}

\vspace{8pt}
\noindent
\textbf{Base Model Output:} \textcolor{red}{$\times$}

Analyzes:
\[
10^j - 10^i = 10^i(10^{j-i} - 1)
\]
Assumes:
\[
10^{j-i} - 1 \text{ divisible by } 1001^2 = 7^2 \cdot 11^2 \cdot 13^2
\]
Finds:
\[
k = j - i \text{ must be even and divisible by } 6
\]
Counts even $k$ from 6 to 98:
\[
\frac{98 - 6}{2} + 1 = \boxed{47}
\]

\vspace{1em}
\textbf{Fine-Tuned Model Output:} \textcolor{green}{$\checkmark$}

Iterates over all even $i < j \leq 99$, checks:
\[
(10^j - 10^i) \bmod 1001^2 = 0
\]

Python verification:
\begin{verbatim}
def count_multiples():
    count = 0
    for i in range(1, 50):
        for j in range(i + 1, 50):
            if (10**(2*j) - 10**(2*i)) % 1001**2 == 0:
                count += 1
    return count

print(count_multiples())
\end{verbatim}

Output: \texttt{0}

Final result: $\boxed{0}$

\vspace{1em}
\textbf{Ground Truth:} $\boxed{0}$
\end{tcolorbox}

\vspace{1em}
\noindent\textbf{Example 3: Rate-Time-Depth Problem}

\begin{tcolorbox}[colback=green!5, colframe=green!50!black, title=Base Model Fails -- Fine-Tuned Model Succeeds, fonttitle=\bfseries, breakable]
\small
\begin{center}
\includegraphics[width=0.55\textwidth]{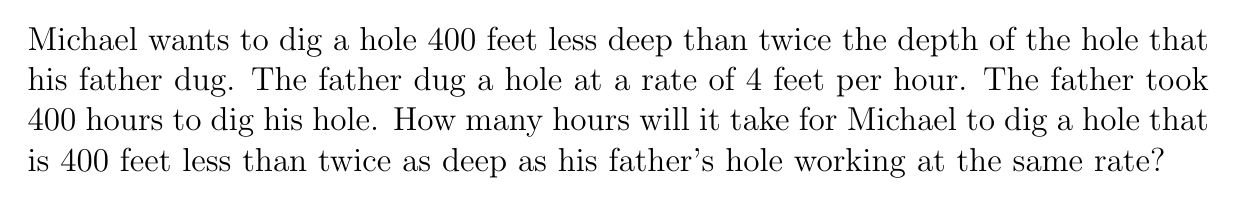}\\[6pt]
\textbf{Problem:} Rate-time-depth problem where the base model fails and the fine-tuned model succeeds.
\end{center}

\vspace{8pt}
\noindent
\textbf{Base Model Output:} \textcolor{red}{$\times$}

To solve this problem:
\begin{enumerate}
    \item Assumes father's hole depth is 400 feet.
    \item Calculates twice the depth: $2 \times 400 = 800$ feet.
    \item Subtracts 400: $800 - 400 = 400$ feet.
    \item Computes time: $400 \div 4 = \boxed{100}$ hours.
\end{enumerate}

\vspace{1em}
\textbf{Fine-Tuned Model Output:} \textcolor{green}{$\checkmark$}

\begin{enumerate}
    \item Father's hole depth: $4 \text{ ft/hr} \times 400 \text{ hr} = 1600$ ft.
    \item Michael's hole depth: $2 \times 1600 - 400 = 2800$ ft.
    \item Time to dig: $2800 \div 4 = 700$ hours.
\end{enumerate}

Python verification:
\begin{verbatim}
def calculate_time():
    rate = 4
    time_father = 400
    depth_father = rate * time_father
    depth_michael = 2 * depth_father - 400
    time_michael = depth_michael / rate
    return time_michael

result = calculate_time()
print(result)
\end{verbatim}

Output: \texttt{700.0}

Final result: $\boxed{700}$

\vspace{1em}
\textbf{Ground Truth Solution:}

Father's hole depth: $4 \text{ ft/hr} \times 400 \text{ hr} = 1600 \text{ ft}$.
Michael's hole depth: $2 \times 1600 - 400 = 2800 \text{ ft}$.
Time to dig: $2800 \div 4 = \boxed{700}$ hours.
\end{tcolorbox}

\vspace{1em}
\noindent\textbf{Example 4: Modular Congruence Problem}

\begin{tcolorbox}[colback=green!5, colframe=green!50!black, title=Base Model Fails -- Fine-Tuned Model Succeeds, fonttitle=\bfseries, breakable]
\small
\begin{center}
\includegraphics[width=0.5\textwidth]{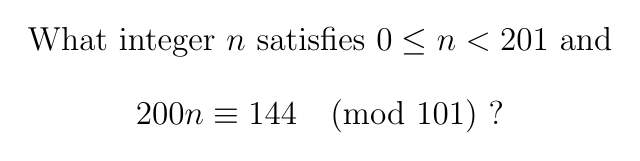}\\[6pt]
\textbf{Problem:} Modular arithmetic problem: $200n \equiv 144 \pmod{101}$
\end{center}

\vspace{8pt}
\noindent
\textbf{Base Model Output:} \textcolor{red}{$\times$}

To solve the congruence $200n \equiv 144 \pmod{101}$, the model simplifies:
\[
200 \equiv -1 \pmod{101} \Rightarrow -n \equiv 144 \pmod{101}
\]
\[
n \equiv -144 \pmod{101} \Rightarrow n \equiv \boxed{58} \pmod{101}
\]

\vspace{1em}
\textbf{Fine-Tuned Model Output:} \textcolor{green}{$\checkmark$}

Simplifies:
\[
200 \equiv 99 \pmod{101} \Rightarrow 99n \equiv 144 \pmod{101}
\]
Finds inverse of $99 \bmod 101$, then solves:
\[
n \equiv 29 \pmod{101}
\]

Python verification:
\begin{verbatim}
from sympy import mod_inverse

def find_n():
    a = 144
    b = 99
    m = 101
    inv_b = mod_inverse(b, m)
    return (a * inv_b) % m

print(find_n())
\end{verbatim}

Output: \texttt{29}

Final result: $\boxed{29}$

\vspace{1em}
\textbf{Ground Truth:} $\boxed{29}$
\end{tcolorbox}

\end{document}